 \documentclass[pmlr,twocolumn,10pt]{jmlr} 

\usepackage{booktabs}
\usepackage{siunitx}

\usepackage[switch]{lineno}

\usepackage[most]{tcolorbox}
\usepackage{adjustbox}
\usepackage{makecell}
\usepackage{multirow}
\usepackage{multicol}
\usepackage{ragged2e}
\usepackage{threeparttable}
\usepackage{caption}

\usepackage[T1]{fontenc}
\usepackage[utf8]{inputenc}

\newcommand{\equal}[1]{{\hypersetup{linkcolor=black}\thanks{#1}}}

\theorembodyfont{\upshape}
\theoremheaderfont{\scshape}
\theorempostheader{:}
\theoremsep{\newline}

\jmlrvolume{297}
\jmlryear{2025}
\jmlrsubmitted{LEAVE UNSET}
\jmlrpublished{LEAVE UNSET}
\jmlrworkshop{Machine Learning for Health (ML4H) 2025} 

 \title[SQL Correction and Question Answerability Classification for Reliable EHR QA]{SCARE: A Benchmark for \underline{S}QL \underline{C}orrection and Question \underline{A}nswerability Classification for \underline{R}eliable \underline{E}HR Question Answering}

\author{%
\Name{Gyubok Lee}\equal{These authors contributed equally}\footnotemark[2] \Email{gyubok.lee@kaist.ac.kr}\\
\Name{Woosog Chay}\footnotemark[1]\footnotemark[2] \Email{benchay@kaist.ac.kr}\\
\Name{Edward Choi}\footnotemark[2] \Email{edwardchoi@kaist.ac.kr}\\
\addr \footnotemark[2] Korea Advanced Institute of Science \& Technology, South Korea \\
\addr \footnotemark[1] Co-first authors \\
}

\begin{document}

\maketitle

\begin{abstract}
Recent advances in Large Language Models (LLMs) have enabled the development of text-to-SQL models that allow clinicians to query structured data stored in Electronic Health Records (EHRs) using natural language. However, deploying these models for EHR question answering (QA) systems in safety-critical clinical environments remains challenging: incorrect SQL queries—whether caused by model errors or problematic user inputs—can undermine clinical decision-making and jeopardize patient care.
While prior work has mainly focused on improving SQL generation accuracy or filtering questions before execution, there is a lack of a unified benchmark for evaluating independent post-hoc verification mechanisms (i.e., a component that inspects and validates the generated SQL before execution), which is crucial for safe deployment.
To fill this gap, we introduce \textsc{SCARE}, a benchmark for evaluating methods that function as a post-hoc safety layer in EHR QA systems. \textsc{SCARE} evaluates the joint task of (1) classifying question answerability (\textit{i.e.}, determining whether a question is answerable, ambiguous, or unanswerable) and (2) verifying or correcting candidate SQL queries. The benchmark comprises 4,200 triples of questions, candidate SQL queries, and expected model outputs, grounded in the MIMIC-III, MIMIC-IV, and eICU databases. 
It covers a diverse set of questions and corresponding candidate SQL queries generated by seven different text-to-SQL models, ensuring a realistic and challenging evaluation.
Using \textsc{SCARE}, we benchmark a range of approaches—from two-stage methods to agentic frameworks. Our experiments reveal a critical trade-off between question classification and SQL error correction, highlighting key challenges and outlining directions for future research.
\end{abstract}


\paragraph*{Data and Code Availability}
Data and code are publicly available on our GitHub repository at \url{https://github.com/glee4810/SCARE}.

\paragraph*{Institutional Review Board (IRB)}
IRB approval is not required for this work. The patient records used in this work are from the PhysioNet website and licensed under the Open Data Commons Open Database License v1.0\footnote{\url{https://opendatacommons.org/licenses/odbl/1-0/}}.


\section{Introduction}
\label{sec:intro}

Electronic Health Records (EHRs) store a wide range of patient data, such as hospital admissions, diagnoses, procedures, and prescriptions, making them essential for healthcare practice and research.
Advances in Large Language Models (LLMs) have enabled natural language interaction with EHRs, with text-to-SQL models converting clinicians' questions into SQL queries to retrieve patient data without requiring SQL expertise \citep{wang2020text, lee2022ehrsql, bardhan2024question}.

However, deploying these systems in clinical settings carries significant risks. For example, an incorrect SQL query could miss a patient's penicillin allergy or miscalculate a medication dosage, leading to potentially catastrophic outcomes.
To ensure safe deployment, question answering (QA) systems over EHRs must generate accurate queries that reflect user intent or reject unsuitable ones. However, achieving such reliability is hindered by two key challenges:
(1) Clinicians, often unfamiliar with SQL or database systems, may pose problematic questions \citep{lee2022ehrsql}. These include \emph{unanswerable} queries (\textit{e.g.}, requesting physician data not in the schema) and \emph{ambiguous} ones (\textit{e.g.}, ``BP?'') that must be clarified rather than translated directly into SQL. (2) Even well-posed (\emph{answerable}) questions can result in incorrect SQL due to limitations in text-to-SQL models \citep{tarbell2023understandinggeneralizationmedicaltexttosql, lee-etal-2024-overview, shi2024ehragent}.

To address these challenges, we introduce \textsc{SCARE}\footnote{A benchmark for \textbf{S}QL \textbf{C}orrection and Question \textbf{A}nswerability Classification for \textbf{R}eliable \textbf{E}HR question answering}, a benchmark designed to evaluate a post-hoc reliability layer in EHR QA systems. 
Unlike existing benchmarks that focus solely on either SQL correction \citep{pourreza2024din, wang2023mac, magic} or question answerability classification \citep{triagesql, dte}, \textsc{SCARE} is the first to evaluate an integrated post-hoc layer that handles both tasks, ensuring more reliable QA for EHRs.
To construct \textsc{SCARE}, we first source QA data compatible with three major EHR databases---MIMIC-III \citep{johnson2016mimic}, MIMIC-IV \citep{johnson2023mimic}, and eICU \citep{pollard2018eicu}---and augment it with manually annotated ambiguous and unanswerable questions to capture a diverse range of problematic user queries.
We then generate SQL queries from these answerable, ambiguous, and unanswerable questions using a diverse set of text-to-SQL models, from lightweight options to advanced agentic frameworks, and pair them with the corresponding answers (expected model outputs), yielding 4,200 question–SQL–answer tuples.
Using \textsc{SCARE} as a testbed, we compare various types of methods. Our experiments reveal significant challenges faced by existing models in handling the integrated task, particularly in balancing the need to correct flawed queries while preserving already-correct ones, and in accurately identifying nuanced ambiguities. These findings highlight the difficulty of robust post-hoc verification and underscore the necessity of \textsc{SCARE} for driving future research toward the safe deployment of clinical QA systems.

\begin{figure*}[t!]
\centering
\includegraphics[width=\textwidth]{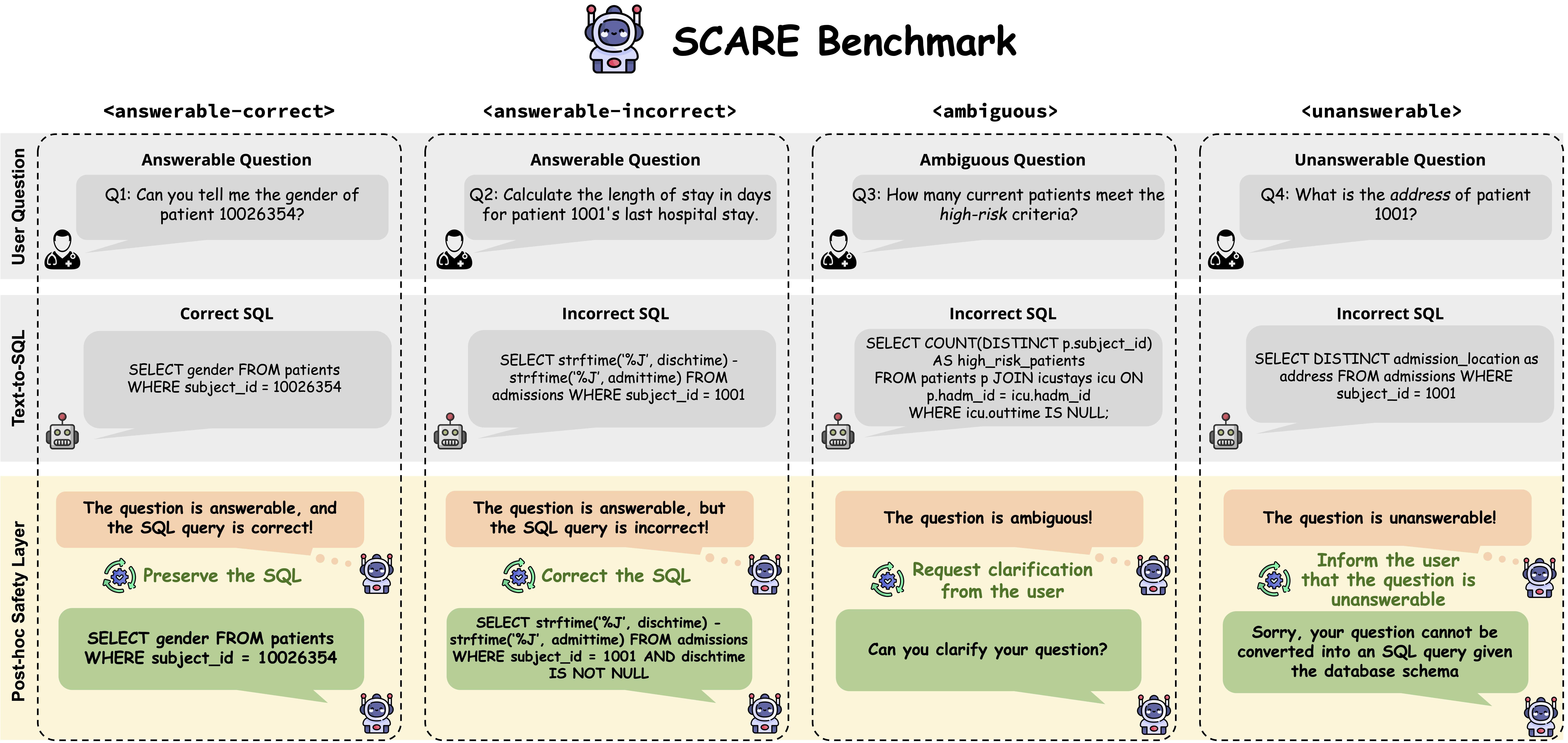}
\caption{Overview of the \textsc{SCARE} benchmark for evaluating a post-hoc verification layer. The task assumes a candidate SQL query has already been generated by an upstream (and potentially black-box) text-to-SQL model. The layer then takes both the user's question and the candidate SQL as input to decide one of four actions: (1) for an answerable question with correct SQL, preserve the query; (2) for an answerable question with incorrect SQL, correct the query; (3) for an ambiguous question, identify the ambiguity for user clarification; and (4) for an unanswerable question, reject the query and inform the user, thus ensuring the overall reliability and safety of the EHR QA system.}
\label{fig:motivation}
\end{figure*}

\section{Related Work}
\subsection{EHR QA for Structured Data}
EHR question answering (QA) involves answering clinically relevant questions by querying patient data stored in EHRs. These tasks may require clinical knowledge or time-based reasoning.
MIMICSQL \citep{wang2020text} and EHRSQL \citep{lee2022ehrsql}, for example, employ text-to-SQL modeling to support question answering over structured EHR databases, demonstrating the potential for querying large-scale datasets using natural language. 
Similarly, \citet{raghavan2021emrkbqa} uses a method that first converts natural language questions into logical forms and then translates them into SQL.
Alternative approaches have used SPARQL \citep{park2021knowledge} or Python-based agents \citep{shi2024ehragent} to perform EHR QA. However, text-to-SQL methods remain the most widely used technology due to their efficiency and scalability for large-scale databases like EHRs.

\subsection{Reliability in Text-to-SQL}

Alongside advances in text-to-SQL models—from few-shot prompting \citep{rajkumar2022evaluating, chang2023prompt} to advanced task decomposition \citep{pourreza2024din} and multi-agent frameworks \citep{wang2023mac}—a stream of research has focused on enabling more reliable deployment of these models, which can be broadly categorized into three areas.

\paragraph{SQL Error Correction.} This line of work enhances reliability by reducing model errors in SQL generation, aiming to improve overall performance. The task typically involves taking an initial SQL query as input and producing a corrected version as output \citep{gong2025sqlensendtoendframeworkerror, magic}. Although vital for boosting accuracy, these approaches assume that all input questions are convertible to SQL (\textit{i.e.}, answerable), limiting their applicability in real-world clinical scenarios where user inputs are unconstrained.

\paragraph{Question Answerability Classification.} A second strand of research focuses exclusively on identifying problematic user questions, typically before SQL generation. Benchmarks like TriageSQL \citep{triagesql} and DTE \citep{dte} focus on classifying questions as answerable or one of several non-answerable categories. Although this tackles an essential layer of input validation, these methods act as pre-hoc filters and do not handle SQL correction, resulting in a gray area where neither task fully covers the combined challenge of question filtering and query verification.

\paragraph{SQL Generation with Abstention.} Most closely related to our goal are works that integrate reliability into the end-to-end generation process. \citet{lee2022ehrsql, lee-etal-2024-overview, lee2024trustsql, somov2025confidence} evaluate models on their ability to generate accurate SQL while abstaining if the query is deemed incorrect, where the concept of incorrect SQL covers both model errors for answerable questions and intrinsic failure due to invalid user input such as ambiguous or unanswerable questions. While these approaches bridge the gap by incorporating caution directly into the pipeline, their evaluation focuses on an all-or-nothing outcome: produce a perfect query or abstain. This conflates the SQL generator's capabilities with the system's verification mechanism, making it difficult to assess their interplay. Furthermore, this all-or-nothing approach limits nuanced user interactions, such as noting that a question is ambiguous or that a request is impossible to fulfill using SQL given the database schema.


\section{Problem Formulation in \textsc{SCARE}}
\label{sec:task_formulation}

We define the task of \textit{post-hoc verification}, performed by an independent safety layer within a safety-critical EHR QA system. This layer audits the output of an upstream text-to-SQL model---treated as a black box for modularity---before execution. The task requires three inputs: a natural language question $q$, the database schema $S$, and a candidate SQL query $\hat{y}$. Crucially, $\hat{y}$ is provided for all question types to reflect a real-world deployment scenario where an upstream model might still produce SQL for flawed inputs. The layer must therefore decide whether to preserve the candidate, correct it, or reject it by informing the user.

The \textsc{SCARE} benchmark contains four categories aligned with Fig.~\ref{fig:motivation}:
\begin{itemize}
    \item \textbf{\texttt{answerable-correct} $(q_{ans}, \hat{y}^{cor}, y^*)$}: $q_{ans}$ is an answerable question and $\hat{y}^{cor}$ is the correct candidate SQL. The correctness is determined by comparing its execution result to that of the ground-truth SQL, $y^*$. 
    \item \textbf{\texttt{answerable-incorrect} $(q_{ans}, \hat{y}^{inc}, y^*)$} : The candidate SQL query is incorrect. 
    \item \textbf{\texttt{ambiguous} $(q_{amb}, \hat{y}^{inc}, l_{amb})$}: $q_{amb}$ is an ambiguous question and the ground-truth label $l_{amb}$ is the string ``ambiguous''.
    \item \textbf{\texttt{unanswerable} $(q_{una}, \hat{y}^{inc}, l_{una})$} : $q_{una}$ is an unanswerable question and the ground-truth label $l_{una}$ is the string ``unanswerable''.
\end{itemize}

\noindent For both \texttt{ambiguous} and \texttt{unanswerable}, any candidate $\hat{y}$ is considered incorrect because these questions either require clarification for accurate SQL translation or lie outside the SQL functionalities given the provided schema.

Given $(q,S,\hat{y})$, the model $f$ is expected to output a verified result $o^*\in\{y^{\prime}, \ell_{\mathrm{amb}}, \ell_{\mathrm{una}}\}$:

\begin{equation}
\small
f(q,S,\hat{y}) = \begin{cases}
{y^{\prime}} &  \text{if $q$ is answerable} \\
l_{amb} &  \text{if $q$ is ambiguous} \\
l_{una}& \text{if $q$ is unanswerable}, \\
\end{cases}
\label{eq:task-formulation}
\end{equation}
\noindent where $y^{\prime}$ is the output SQL (either the preserved $\hat{y}$ or a corrected SQL query) such that its execution result matches that of the ground-truth SQL $y^{*}$.


\begin{figure}[t!]
\centering
\includegraphics[width=8cm]{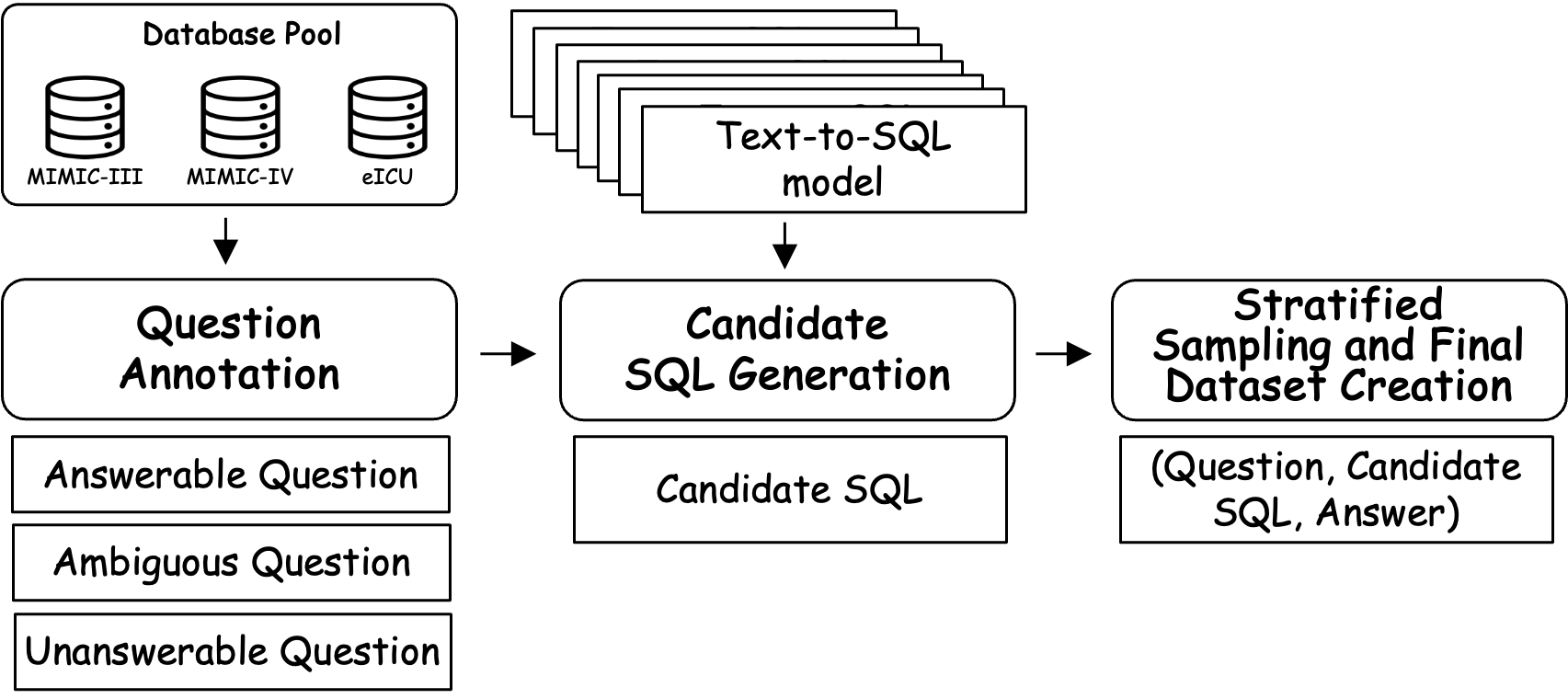}
\caption{Overview of the \textsc{SCARE} benchmark construction pipeline.}
\label{fig:data_creation_pipeline}
\end{figure}

\section{Benchmark Construction}

This section describes the construction process of the \textsc{SCARE} benchmark in three main stages: (1) annotating a diverse set of questions, including answerable, ambiguous, and unanswerable ones, across three EHR databases; (2) generating candidate SQL queries using multiple text-to-SQL models; and (3) conducting stratified sampling to create a balanced evaluation set. An overview of the process is shown in Figure~\ref{fig:data_creation_pipeline}.

\subsection{Question Annotation}
\label{sec:question_curation}

\newcolumntype{A}{>{\RaggedRight\ttfamily}p{0.10\textwidth}}
\newcolumntype{B}{>{\RaggedRight}p{0.20\textwidth}}
\newcolumntype{C}{>{\RaggedRight\ttfamily}p{0.34\textwidth}}
\newcolumntype{D}{>{\RaggedRight\ttfamily}p{0.24\textwidth}}

\begin{table*}[t!]
\centering
\footnotesize
\renewcommand{\arraystretch}{1.0}
\caption{Sample data from the \textsc{SCARE} benchmark. For \texttt{answerable-correct}, the model is expected to output the same SQL, as it is correct. For \texttt{answerable-incorrect}, the candidate SQL is flawed because it fails to use \texttt{DISTINCT} when counting patients, so the model is expected to correct the error in the SQL. For \texttt{ambiguous}, the model is expected to output ``ambiguous,'' as the question contains the vague word ``enough.'' For \texttt{unanswerable}, the model is expected to output ``unanswerable,'' as ``family visitation'' goes beyond the information stored in the schema.}
\label{tab:scare_sample_final}
\begin{tabular}{@{} A B C D @{}}
\toprule
\multicolumn{1}{c}{\textbf{Type}} & 
\multicolumn{1}{c}{\textbf{User Question}} & 
\multicolumn{1}{c}{\textbf{Candidate SQL}} & 
\multicolumn{1}{c}{\textbf{Answer}} \\
\midrule
answerable- correct &
Count the ICU visits of patient 10007058 since 2100. &
SELECT COUNT(*) FROM icustays WHERE subject\_id = 10007058 AND intime >= `2100-01-01 00:00:00' AND intime <= `2100-12-31 23:59:00' &
SELECT COUNT(*) FROM icustays WHERE subject\_id = 10007058 AND intime >= `2100-01-01 00:00:00' AND intime <= `2100-12-31 23:59:00' \\
\midrule
answerable- incorrect &
How many people were admitted to the hospital? &
SELECT count(subject\_id) FROM admissions &
SELECT count(DISTINCT subject\_id) FROM admissions \\
\midrule
ambiguous &
How many patients were administered divalproex (delayed release) in \textit{enough} doses since 2100? &
SELECT count(DISTINCT subject\_id) FROM prescriptions WHERE drug = `divalproex' AND stoptime > `2100-01-01' &
`ambiguous' \\
\midrule
unanswerable &
Did patient 10007795 have \textit{family visitation} during their first ICU stay? &
SELECT ce.charttime AS visitation\_time FROM chartevents ce JOIN d\_items di ON ce.itemid = di.itemid JOIN icustays icu ... &
`unanswerable' \\
\bottomrule
\end{tabular}
\vspace{-2mm}
\end{table*}

\paragraph{EHR Databases.}
Our work uses three major publicly available EHR databases: MIMIC-III \citep{johnson2016mimic}, MIMIC-IV \citep{johnson2023mimic}, and eICU \citep{pollard2018eicu}. For MIMIC-III, we adopt the schema from the MIMICSQL \citep{wang2020text} dataset. For MIMIC-IV and eICU, we follow the preprocessing procedure used in EHRSQL \citep{lee2022ehrsql}. The resulting schema sizes are: MIMIC-III (5 tables, 50 columns), MIMIC-IV (7 tables, 112 columns), and eICU (10 tables, 72 columns).\footnote{We use the demo versions of MIMIC-IV and eICU to avoid privacy alterations such as value shuffling, which perturbs patient data distributions. The demos share the same database schema structures as the full versions.}. Using these databases as the foundation for our benchmark, we create a pool of answerable questions, followed by unanswerable and ambiguous ones.

\subsubsection{Answerable Question Creation}

For questions compatible with MIMIC-III, we use 1,000 question–SQL pairs from MIMICSQL (\texttt{mimicsql\_natural\_v2}). For MIMIC-IV and eICU, however, we cannot directly reuse the EHRSQL pairs, because their value-shuffled patient data do not match our databases. Instead, we construct new data from scratch using the EHRSQL question templates (\textit{e.g.}, ``What is the route of administration of \{drug\_name\}?'' where the actual value for \{drug\_name\} is later sampled from the database). The construction proceeds in three steps: (1) sample valid values from the databases, (2) generate new ground-truth SQL queries, and (3) paraphrase the questions using OpenAI's GPT-4o. This process results in 450 pairs for MIMIC-IV and 432 pairs for eICU. These SQL queries are considered the ground-truth answers for the answerable portion of our dataset (\textbf{\texttt{answerable-correct}} and \textbf{\texttt{answerable-incorrect}}).

\subsubsection{Ambiguous and Unanswerable Question Annotation}
\label{sec:ambig_unans_anno}

During the deployment of EHR QA systems, users often pose unanswerable or ambiguous questions for which no corresponding SQL exists. To address this issue, we curate such questions by annotating new instances into six categories: three types of ambiguity and three types of unanswerability, derived from prior text-to-SQL literature and from frequently unanswerable cases included in EHRSQL \cite{lee2022ehrsql}. To ensure a diverse and comprehensive set of problematic questions, newly annotated questions are first generated using GPT-4o based on their target categories, followed by human validation to ensure proper alignment between each question and its assigned category, while filtering out semantically similar queries. Descriptions of these categories are provided below, with further annotation details and examples in Appendix~\ref{apd:question_curation_annotation}.

\paragraph{Ambiguous Questions.} 
Ambiguous questions are those that require clarification before SQL generation (the model answer to these questions is  ``ambiguous'').  These include \texttt{vague-question} (\textbf{VQ}) instances, such as short, phrasal questions (\textit{e.g.}, ``Patient status?'') \citep{radhakrishnan2020colloql, dte}; \texttt{vague-word} (\textbf{VW}) instances with imprecise terms (\textit{e.g.}, ``Recent high blood pressure cases'') \citep{he2024text2analysis}; and \texttt{ambiguous-reference} (\textbf{AR}) instances involving unclear entities (\textit{e.g.}, ``The patient from last week'') \citep{yu2019cosql, zhu2024large}.
For reliable EHR QA systems, it is crucial to classify such questions as ``ambiguous'' and notify users that further clarification is needed to ensure accurate query processing.

\paragraph{Unanswerable Questions.}
Unanswerable questions cannot be resolved via SQL queries given the provided database schema (the model answer to these questions is ``unanswerable'').
These include \texttt{small-talk} (\textbf{ST}), encompassing casual talk unrelated to EHR (\textit{e.g.}, ``What's the weather like today?'') \citep{triagesql}; \texttt{out-of-scope} (\textbf{OS}), where requests go beyond SQL capabilities (\textit{e.g.}, ``Predict future patient outcomes'') \citep{triagesql, dte}; and \texttt{missing-column}, including newly created examples (\textbf{MC$^{\prime}$}) and those adapted from EHRSQL (\textbf{MC$^{\prime\prime}$}), where questions reference non-existent columns (\textit{e.g.}, ``Query the blood\_type column,'' but it doesn't exist) \citep{triagesql, dte, lee2022ehrsql}. 
For reliable EHR QA systems, it is crucial to classify such questions as ``unanswerable'' and inform users that the request cannot be fulfilled within the scope of the text-to-SQL task. This notification is important for guiding users to refine their next input to align with the system's capabilities.

\subsection{Candidate SQL Generation}
\label{sec:candidate_sql_generation}

Building on the questions, we generate candidate SQL queries to simulate real-world model outputs that could be implemented within the EHR QA system.

\subsubsection{Generating Candidate SQL}

To generate a diverse pool of candidate SQL queries ($\hat{y}$), we utilize a variety of text-to-SQL models, ranging from fine-tuned SQL-specialized models to agentic frameworks. This diverse model set enables stress-testing of error-correction frameworks against varied $\hat{y}$ given $q$ distributions, accommodating diverse question types such as answerable, ambiguous, and unanswerable queries for EHR QA.
The models include: \textbf{\textsc{LLM-SQL}}, a few-shot baseline utilizing GPT-4o for SQL generation \citep{chang2023prompt}; \textbf{\textsc{CodeS-15B}}, a 15B parameter model pre-trained for SQL and fine-tuned on the BIRD dataset \citep{li2024codes}; \textbf{\textsc{DIN-SQL}}, an advanced in-context learning approach using GPT-4o with task decomposition and self-correction \citep{pourreza2024din}; \textbf{\textsc{MAC-SQL}}, a multi-agent framework employing GPT-4o for iterative SQL query refinement \citep{wang2023mac}; \textbf{\textsc{Deepseek R1-70B}}, a 70B parameter general-purpose reasoning model \citep{guo2025deepseek}; \textbf{\textsc{o4-mini}}, an advanced reasoning model from OpenAI; and \textbf{\textsc{Qwen3-32B}}, a 32B parameter general-purpose reasoning model \citep{yang2025qwen3technicalreport}. Table~\ref{tab:sql_generator} reports the execution accuracy of these models.

\begin{table}[t!]
\centering
\huge
\renewcommand{\arraystretch}{1.5}
\caption{The dataset statistics of the \textsc{SCARE} benchmark.}
\begin{adjustbox}{width=\linewidth,center}
\begin{tabular}{cccccc}
\toprule
\textbf{Database }& \makecell{$\texttt{answerable-}$ \\ $\texttt{correct}$} & \makecell{$\texttt{answerable-}$ \\ $\texttt{incorrect}$} & $\texttt{ambiguous}$ & $\texttt{unanswerable}$ & \textbf{Total} \\
\midrule
\textbf{MIMIC-III} & 350  & 350  & 350 & 350 & 1,400 \\
\textbf{MIMIC-IV} & 350  & 350  & 350 & 350 & 1,400 \\
\textbf{eICU} & 350  & 350  & 350 & 350 & 1,400 \\
\midrule
\textbf{Tota}l & 1,050 & 1,050 & 1,050 & 1,050 & 4,200 \\
\bottomrule
\end{tabular}
\end{adjustbox}
\label{tab:data_statistics}
\end{table}

\subsection{Stratified Sampling and Final Dataset Creation}
Based on the pool of generated SQL candidates, we construct the final benchmark through stratified sampling, dividing the data into subgroups (strata) to ensure a balanced distribution across different scenarios. For each of the three databases, we select 1,400 instances, evenly balanced across four strata (350 instances each):
\texttt{answerable-correct}, where the candidate SQL returns the correct answer, matching the ground-truth result;
\texttt{answerable-incorrect}, where the candidate SQL returns an incorrect answer, often due to issues like wrong column selection or invalid operations; 
\texttt{ambiguous}, where instances are derived from ambiguous questions, making any candidate SQL inherently invalid; and
\texttt{unanswerable}, where instances are derived from unanswerable questions, making any candidate SQL inherently invalid. 
After sampling from each database independently, we combine these selections across the three databases to form the complete benchmark, resulting in a well-balanced dataset of 4,200 instances overall (3 databases $\times$ 1,400 instances each). Rather than reflecting a naturally skewed real-world distribution of user queries, this stratified, balanced split ensures that each of the four key safety scenarios across different EHR databases is equally tested, which is essential for the diagnostic evaluation of post-hoc verification models in EHR QA systems. Sample data and data statistics are shown in Tables~\ref{tab:scare_sample_final} and~\ref{tab:data_statistics}.

\section{Experiments}
\subsection{Metrics}
\label{sec:metrics}

For the two \textbf{\texttt{answerable}} categories, we use three metrics. First, we measure \textbf{Coverage (Cov)}, the proportion of instances where the model provides a SQL output rather than a classification label (\textit{i.e.}, ``ambiguous'', or ``unanswerable''). Then, we measure the \textbf{Preservation Rate (PR)} on \texttt{answerable-correct} inputs and the \textbf{Correction Rate (CR)} on \texttt{answerable-incorrect} inputs. Both metrics are defined as the final execution accuracy, calculated over the total number of instances in their respective categories. This ensures that models are penalized not only for errors in SQL generation but also for incorrectly classifying an answerable question as \texttt{ambiguous} or \texttt{unanswerable}.

For the \textbf{\texttt{unanswerable}} and \textbf{\texttt{ambiguous}} categories, we evaluate the model's ability to correctly classify questions into their respective categories. We report the per-class \textbf{Precision}, \textbf{Recall}, and \textbf{F1-score} for both the ``unanswerable'' and ``ambiguous'' labels.

\begin{table*}[t!]
\centering
\footnotesize
\caption{Baseline performance on the \textsc{SCARE} benchmark. Methods classify questions as answerable, ambiguous, or unanswerable, and handle SQL queries for answerable questions by preserving correct queries or fixing incorrect ones. Metrics are defined in Section \ref{sec:metrics}. Higher values indicate better performance.}

\renewcommand{\arraystretch}{1.3}
\begin{adjustbox}{width=\linewidth,center}
\begin{tabular}{ccccccccccc}
\toprule

\multirow{3}{*}{\makecell{\textbf{Method}}}
& \multicolumn{2}{c}{\makecell{\textbf{\texttt{answerable-}} \\ \textbf{\texttt{correct}}}}
& \multicolumn{2}{c}{\makecell{\textbf{\texttt{answerable-}} \\ \textbf{\texttt{incorrect}}}}
& \multicolumn{3}{c}{\textbf{\texttt{ambiguous}}}
& \multicolumn{3}{c}{\textbf{\texttt{unanswerable}}}\\

\cmidrule(lr){2-3}
\cmidrule(lr){4-5}
\cmidrule(lr){6-8}
\cmidrule(lr){9-11}

& \textbf{$\mathbf{PR}$} 
& \textbf{$\mathbf{Cov}$}
& \textbf{$\mathbf{CR}$}
& \textbf{$\mathbf{Cov}$}
& \textbf{$\mathbf{Prec}$}
& \textbf{$\mathbf{Rec}$} 
& \textbf{$\mathbf{F1}$}
& \textbf{$\mathbf{Prec}$}
& \textbf{$\mathbf{Rec}$} 
& \textbf{$\mathbf{F1}$}
\\

\midrule
\textsc{Two-Stage} & 80.4 & 81.3 & 42.5 & 77.7 & 72.5 & 36.4 & 48.4 & 58.2 & \textbf{93.0} & 71.6 \\
\textsc{Single-Turn} & \underline{97.9} & \textbf{99.5} & 49.6 & \underline{97.6} & 84.5 & 31.6 & 46.0 & \textbf{84.1} & 70.7 & 76.8 \\
\textsc{Single-Turn-Veri} & \textbf{98.1} & \textbf{99.5} & 52.1 & \textbf{98.1} & \underline{89.0} & 32.4 & 47.5 & \underline{84.0} & 72.5 & 77.8 \\
\textsc{Multi-Turn-SelfRef} & 97.2 & \underline{99.4} & 51.4 & \textbf{98.1} & \textbf{89.4} & 32.9 & 48.1 & 83.5 & 72.7 & 77.7 \\
\textsc{Single-Turn-Cls} & 95.5 & 97.6 & \underline{53.0} & 94.2 & 87.3 & \textbf{39.1} & \textbf{54.0} & 75.4 & 86.6 & \underline{80.6} \\
\textsc{Single-Turn-Veri-Cls} & 95.9 & 97.8 & \textbf{53.8} & 94.0 & 87.0 & \underline{39.0} & \underline{53.9} & 75.7 & \underline{86.8} & \textbf{80.8} \\
\textsc{Multi-Turn-SelfRef-Cls} & 97.8 & 99.4 & \underline{53.0} & 98.1 & 88.0 & 32.9 & 47.9 & 83.3 & 72.7 & 77.6 \\
\bottomrule

\end{tabular}
\end{adjustbox}
\label{tab:main_result}
\end{table*}

\subsection{Methods}

To establish comprehensive baselines on \textsc{SCARE}, we evaluate seven methods that represent distinct strategies for the joint task of question answerability classification and SQL correction. These include four base methods and three hybrid variants built upon them.

\begin{itemize}
    \item \textsc{Two-Stage}: This method follows a modular, divide-and-conquer strategy. It first employs a dedicated classifier to determine if a question is answerable, ambiguous, or unanswerable. Only if the question is deemed answerable does it proceed to a second stage, where a separate module verifies the candidate SQL and corrects it if necessary.
    \item \textsc{Single-Turn}: This method adopts an integrated, end-to-end approach. In a single generative pass, the model is tasked with jointly analyzing the question and candidate SQL to simultaneously handle answerability classification and SQL correction, directly outputting either a final SQL query or a classification label.
    \item \textsc{Single-Turn-Veri}: Building on the end-to-end approach, this method introduces a simple iterative verification loop. It extends \textsc{Single-Turn} by having an internal verifier to check the generated output. If the verifier flags an error, the model performs multiple retries to generate a correct answer.
    \item \textsc{Multi-Turn-SelfRef}: This method involves an iterative refinement strategy in a multi-turn setting. It generates an initial answer, produces feedback on its own output, and then uses this feedback to guide the next refinement attempt.
\end{itemize}

The three variant methods—\textsc{Single-Turn-Cls}, \textsc{Single-Turn-Veri-Cls}, and \textsc{Multi-Turn-SelfRef-Cls}—are designed as hybrid approaches. These aim to combine the strengths of the modular and integrated strategies by using the output from the specialized classifier in \textsc{Two-Stage} (i.e., the classification result and its reasoning) as an explicit guiding signal for the integrated models. We provide the detailed prompts used to implement the four base methods in Appendix~\ref{appendix:prompt}.

For the backbone LLMs, we use Gemini-2.5-Flash \citep{comanici2025gemini25pushingfrontier} as the main LLM throughout the main experiments. Additional results for two other closed-source models (GPT-5 mini \citep{OpenaiGPT5} and Gemini-2.0-Flash \citep{Google20GeminiExpands}) and two open-source models (Llama-3.3-70B \citep{llama33} and Qwen3-32B \citep{yang2025qwen3technicalreport}) are provided in Appendix~\ref{appendix:full_results}, all of which exhibit similar performance trends when used to implement the methods.

\subsection{Results}

Table \ref{tab:main_result} shows the performance of various baseline methods on the \textsc{SCARE} benchmark.

\paragraph{Trade-off between question classification and SQL correction.} A critical challenge lies in balancing the need to preserve correct SQL queries (PR) while accurately identifying ambiguous or unanswerable questions. The \textsc{Two-Stage} approach, which decouples these tasks, achieves the highest recall for \texttt{unanswerable} (93.0\%). However, this comes at a significant cost to Cov, as the initial classification stage frequently misidentifies answerable questions. Conversely, \textsc{Single-Turn} excels at PR (97.9\%) by integrating the question classification and SQL correction tasks, but its ability to detect ambiguity is notably weak (Recall 31.6\%). This suggests an inherent tension where maximizing preservation often leads to overlooking problematic inputs, and vice versa. We provide a qualitative analysis of the error cases in Section \ref{appendix:qualitative_analysis}.

\paragraph{Iterative refinement improves correction, but limitations remain.} Iterative approaches (\textsc{Single-Turn-Veri} and \textsc{Multi-Turn-SelfRef}) demonstrate improvements in CR without compromising PR. \textsc{Multi-Turn-SelfRef} achieves a CR of 51.4\%, compared to 49.6\% for the basic \textsc{Single-Turn}, while maintaining a high PR (97.2\%). However, the overall correction capability is still limited. A detailed breakdown of correction outcomes by SQL error type in Table \ref{tab:incorrect_sql_ans} (with explanations provided in Section \ref{appendix:sql_error_types}) shows that, although the methods perform reasonably well on localized errors such as Table/Column (T/C) references (55.1\% CR), they struggle considerably with Other Global (OG) errors, which require substantial structural changes to SQL queries (30.8\% CR).

\paragraph{Hybrid approaches yield the best balance.} The most effective strategies leverage the strengths of both decoupled classification and integrated refinement. The \textsc{-Cls} variants, which incorporate the reasoning output from the \textsc{Two-Stage} classifier, significantly enhance overall performance. Notably, \textsc{Single-Turn-Veri-Cls} achieves the highest CR (53.8\%) and the best F1 score for \texttt{unanswerable} (80.8\%), while maintaining a strong PR (95.9\%). This hybrid approach effectively mitigates the trade-offs observed in the base methods, pointing towards the necessity of integrating explicit answerability reasoning into the verification process.

\begin{table}[t!]
\centering
\footnotesize
\caption{Detailed correction rates by SQL error types for \texttt{answerable-incorrect}. T/C, J/G, PV, OL, and OG denote table/column reference errors, JOIN/GROUP BY errors, predicate value errors, other local errors, and other global errors, respectively.}
\renewcommand{\arraystretch}{1.7}
\begin{adjustbox}{width=\linewidth,center}
\begin{tabular}{cccccc}

\toprule



\textbf{CR} &
\textbf{T/C} &
\textbf{J/G} &
\textbf{PV} &
\textbf{OL} &
\textbf{OG} \\

\midrule

\textsc{Two-Stage} & 45.4 & 48.5 & \underline{39.7} & 31.4 & 17.8 \\
\textsc{Single-Turn} & \underline{52.4} & 54.4 & \textbf{50.7} & 40.4 & 28.0 \\
\textsc{Single-Turn-Veri} & \textbf{55.1} & \underline{56.4} & \textbf{50.7} & \underline{41.0} & \underline{29.0} \\
\textsc{Multi-Turn-SelfRef} & \textbf{55.1} & \textbf{58.8} & \textbf{50.7} & \textbf{42.0} & \textbf{30.8} \\

\bottomrule

\end{tabular}
\end{adjustbox}
\label{tab:incorrect_sql_ans}
\end{table}

\paragraph{Nuanced ambiguity remains highly challenging.} Methods consistently struggle to identify \texttt{ambiguous}, achieving a maximum F1 score of only 54.0\%. As detailed in Table \ref{tab:incorrect_sql_unans}, detection rates for \texttt{vague-question} (VQ) and \texttt{vague-word} (VW) are particularly poor, often remaining below 35\%. This indicates that while models can easily identify overt issues like \texttt{small-talk} (ST, $>$94\% recall), they lack the sensitivity required to detect subtle linguistic ambiguities. This deficiency poses a significant risk, as undetected ambiguities can lead to the execution of incorrect SQL queries.


\begin{table}[t!]

\centering
\huge
\caption{Recall for correctly identifying ambiguous and unanswerable questions by granular question categories. See Section \ref{sec:ambig_unans_anno} for question type definitions.}
\renewcommand{\arraystretch}{1.37}
\begin{adjustbox}{width=\linewidth,center}
\begin{tabular}{cccccccc}
\toprule
\textbf{Rec}
& \textbf{VQ}
& \textbf{VW}
& \textbf{AR}
& \textbf{ST}
& \textbf{OS}
& \textbf{MC$^{\prime}$}
& \textbf{MC$^{\prime\prime}$} \\

\midrule
\textsc{Two-Stage} & 30.1 & 22.1 & \textbf{56.9} & \textbf{100.0} & \textbf{99.2} & \textbf{79.1} & \textbf{94.7} \\
\textsc{Single-Turn} & 32.6 & 29.1 & 33.1 & \underline{98.0} & 74.1 & 53.6 & 60.3 \\
\textsc{Single-Turn-Veri} & 33.7 & 29.4 & 34.0 & \underline{98.0} & 77.0 & 54.0 & 64.0 \\
\textsc{Multi-Turn-SelfRef} & \underline{34.8} & 29.9 & 33.7 & \underline{98.0} & 77.0 & 54.8 & 64.0 \\
\textsc{Single-Turn-Cls} & 32.0 & \textbf{32.3} & 53.1 & 95.6 & \underline{92.1} & 72.6 & 87.0 \\
\textsc{Single-Turn-Veri-Cls} & 32.6 & 30.5 & \underline{54.0} & 95.6 & 91.2 & 74.1 & 87.0 \\
\textsc{Multi-Turn-SelfRef-Cls} & \textbf{37.4} & \underline{31.7} & 52.3 & 94.8 & 91.6 & \underline{74.5} & \underline{87.7} \\

\bottomrule
\end{tabular}
\end{adjustbox}
\label{tab:incorrect_sql_unans}
\end{table}

\subsection{Qualitative Analysis of Failure Modes}

To better understand the challenges highlighted by our benchmark, further qualitative analysis of model failures is provided in Appendix \ref{appendix:qualitative_analysis}. As suggested by quantitative results, the most salient errors fall into two categories: failing to detect nuanced ambiguity and failing to correct global SQL errors.

\paragraph{Failures in \texttt{ambiguous} Classification.}
Models consistently fail to identify ambiguity when a question contains vague expressions (e.g., \texttt{vague-word}, \texttt{vague-question}) and the candidate SQL either ignores or misinterprets them. For example, models often incorrectly approve a candidate SQL query for a question like ``Find patients with \textit{sufficient data}'' because the query appears syntactically valid, failing to recognize that ``sufficient data'' is unresolvable without clarification. In other cases, for phrasal questions like ``Sodium?'', models incorrectly approve a degenerate candidate SQL such as SELECT label FROM d\_labitems WHERE label = `Sodium' instead of classifying the question as ambiguous. These failures persist even with frontier LLMs. 
We hypothesize that this is due to the lack of joint consideration of linguistic vagueness handling and SQL generation when building LLMs. As a result, the models are biased towards generating any SQL that seems fit, rather than assessing the semantic answerability of the question itself.

\paragraph{Failures in \texttt{answerable-incorrect} Correction.}
For answerable questions, models struggle most with Other Global (OG) errors, which require substantial logical or structural corrections to the candidate SQL. These failures mostly stem from two main issues: limited ability to follow instructions and overreliance on parametric knowledge. First, models violate explicit textual guidelines provided in the prompt. For example, when instructed to use the earliest diagnosis time if multiple records exist, a model often overlooks this instruction and preserves incorrect SQL candidate queries. Second, models hallucinate SQL logic based on their internal knowledge instead of using the database schema provided in the context window. These include cases where models invent non-existent ICD codes or table relationships. These cases show the limitations of current LLMs in strictly following complex instructions and leveraging provided context over internal parametric knowledge.

\section{Conclusion}

The safe deployment of EHR question answering systems in clinical environments demands reliability mechanisms that go beyond standard text-to-SQL generation accuracy. In this work, we introduce \textsc{SCARE}, the first benchmark specifically designed to evaluate a unified post-hoc safety layer tasked with the joint challenge of SQL correction and question answerability classification. Grounded in open-source EHR databases, \textsc{SCARE} incorporates diverse scenarios derived from various text-to-SQL models.

Our comprehensive evaluation reveals critical limitations in current approaches. We uncover a stark trade-off between preserving correct queries and accurately identifying problematic questions (either ambiguous or unanswerable). Furthermore, our experimental results reveal that methods severely struggle to detect nuanced ambiguities commonly posed during EHR QA. While hybrid approaches combining iterative refinement with explicit classification signals show promise, significant advancements are still needed before these systems can be reliably deployed in safety-critical clinical applications. \textsc{SCARE} provides an essential tool for the community to drive future research toward developing robust and auditable verification methods, ultimately facilitating the safe integration of LLMs into clinical workflows.

\section{Limitations and Future Work}
The SCARE benchmark has several limitations rooted in its specific design choices. First, its balanced data distribution is intentionally designed for a diagnostic stress test, ensuring each of the four key scenarios is evenly evaluated. However, this design in turn does not reflect the natural distribution of queries during real-world clinical deployment. Similarly, the benchmark is grounded in academic EHR schemas (MIMIC-III, MIMIC-IV, and eICU), which are smaller than many production systems. This choice was made to foster transparent and reproducible research, though generalization to larger, proprietary schemas remains an important future challenge. Finally, SCARE evaluates user-system interactions in a single-turn setting. This design enables a controlled evaluation of a model's ability to detect various problematic question-candidate SQL pairs without the confounding effects of dialogue history. We recognize that the resolution of ambiguity, as opposed to its mere detection, is often best handled through multi-turn interaction. Future research can extend the SCARE framework to the actual production-level EHR systems and a multi-turn setting.


\acks{
This work was supported by the Institute of Information \& Communications Technology Planning \& Evaluation (IITP) grants (No.RS-2019-II190075, No.RS-2025-02304967) and National Research Foundation of Korea (NRF) grants (NRF-2020H1D3A2A03100945), funded by the Korea government (MSIT).
}

\bibliography{jmlr-sample}

\clearpage
\appendix

\newtcolorbox[auto counter,
    list inside=prompt, 
    ]{promptbox}[2][]{%
    breakable,
    colback=white, 
    colframe=black, 
    boxrule=0.5pt, 
    arc=0mm, 
    fonttitle=\bfseries,
    title={Prompt \thetcbcounter: #2}, 
    label={#1}, 
    left=5pt, right=5pt, top=2pt, bottom=2pt
}

\section{Full Performance Results}
\label{apd:full_results}
\label{appendix:full_results}

Table \ref{tab:full_main_result} reports the full results for GPT-5 mini, Gemini-2.0-Flash, Llama-3.3-70B, and Qwen3-32B.

\begin{table*}[t!]
\centering
\footnotesize
\caption{Full baseline performance on the \textsc{SCARE} benchmark. Higher values indicate better performance.}
\renewcommand{\arraystretch}{1.4}
\begin{adjustbox}{width=\linewidth,center}
\begin{tabular}{ccccccccccc}
\toprule
\multirow{3}{*}{\makecell{\textbf{Method}}}
& \multicolumn{2}{c}{\makecell{\textbf{\texttt{answerable-}} \\ \textbf{\texttt{correct}}}}
& \multicolumn{2}{c}{\makecell{\textbf{\texttt{answerable-}} \\ \textbf{\texttt{incorrect}}}}
& \multicolumn{3}{c}{\textbf{\texttt{ambiguous}}}
& \multicolumn{3}{c}{\textbf{\texttt{unanswerable}}}\\
\cmidrule(lr){2-3}
\cmidrule(lr){4-5}
\cmidrule(lr){6-8}
\cmidrule(lr){9-11}
& \textbf{$\mathbf{PR}$}
& \textbf{$\mathbf{Cov}$}
& \textbf{$\mathbf{CR}$}
& \textbf{$\mathbf{Cov}$}
& \textbf{$\mathbf{Prec}$}
& \textbf{$\mathbf{Rec}$}
& \textbf{$\mathbf{F1}$}
& \textbf{$\mathbf{Prec}$}
& \textbf{$\mathbf{Rec}$}
& \textbf{$\mathbf{F1}$}
\\
\midrule
\multicolumn{11}{c}{\textit{GPT-5 mini}} \\ 
\textsc{Two-Stage} & 82.4 & 85.6 & 55.1 & 83.9 & 68.1 & 73.7 & 70.8 & 77.7 & 88.4 & 82.7 \\
\textsc{Single-Turn} & 95.7 & 99.3 & 62.6 & 98.4 & 75.5 & 44.3 & 55.8 & 87.7 & 64.6 & 74.4 \\
\textsc{Single-Turn-Veri} & 96.0 & 99.4 & 65.0 & 99.1 & 80.8 & 46.0 & 58.6 & 85.5 & 70.7 & 77.4\\
\textsc{Multi-Turn-SelfRef} &95.7 & 99.5 & 67.8 & 98.7 & 81.8 & 46.1 & 59.0 & 83.5 & 72.3 & 77.5 \\
\textsc{Single-Turn-Cls} & 88.4 & 91.7 & 59.0 & 86.6 & 74.6 & 75.8 & 75.2 & 79.0 & 88.0 & 83.2\\
\textsc{Single-Turn-Veri-Cls} & 88.8 & 92.7 & 63.7 & 90.1 & 77.5 & 75.5 & 76.5 & 80.2 & 88.1 & 84.0\\
\textsc{Multi-Turn-SelfRef-Cls} & 93.2 & 97.9 & 68.1 & 95.9 & 84.9 & 68.5 & 75.8 & 79.6 & 89.0 & 84.0 \\
\midrule
\multicolumn{11}{c}{\textit{Gemini-2.0-Flash}} \\ 
\textsc{Two-Stage} & 78.6 & 80.0 & 27.5 & 75.5 & 77.3 & 35.6 & 48.8 & 54.3 & 95.7 & 69.3 \\
\textsc{Single-Turn} & 98.8 & 99.5 & 29.4 & 94.7 & 69.2 & 13.2 & 22.2 & 78.5 & 68.5 & 73.1 \\
\textsc{Single-Turn-Veri} & 98.9 & 99.4 & 30.7 & 94.2 & 68.6 & 15.0 & 24.6 & 78.7 & 69.8 & 74.0 \\
\textsc{Multi-Turn-SelfRef} & 97.7 & 98.8 & 34.2 & 91.5 & 68.8 & 18.9 & 29.6 & 74.1 & 76.3 & 75.2 \\
\textsc{Single-Turn-Cls} & 96.3 & 97.6 & 30.1 & 89.9 & 84.9 & 33.2 & 47.8 & 70.6 & 92.6 & 80.1 \\
\textsc{Single-Turn-Veri-Cls} & 96.3 & 97.7 & 31.1 & 89.1 & 85.8 & 34.0 & 48.7 & 70.4 & 92.9 & 80.1 \\
\textsc{Multi-Turn-SelfRef-Cls} & 95.0 & 96.4 & 34.0 & 87.1 & 84.8 & 35.5 & 50.1 & 67.7 & 93.2 & 78.4 \\
\midrule
\multicolumn{11}{c}{\textit{Llama-3.3-70B}} \\ 
\textsc{Two-Stage} & 69.0 & 73.1 & 29.9 & 68.1 & 71.4 & 47.6 & 57.1 & 52.9 & 99.0 & 68.9 \\
\textsc{Single-Turn} & 60.4 & 98.5 & 20.8 & 90.4 & 48.8 & 18.6 & 26.9 & 75.6 & 69.6 & 72.5\\
\textsc{Single-Turn-Veri} & 97.1 & 99.2 & 38.1 & 92.3 & 57.6 & 25.3 & 35.2 & 74.5 & 76.5 & 75.5 \\
\textsc{Multi-Turn-SelfRef} &95.1 & 97.4 & 38.9 & 88.1 & 60.0 & 31.8 & 41.6 & 72.0 & 77.6 & 74.7 \\
\textsc{Single-Turn-Cls} & 74.1 & 77.0 & 29.5 & 69.6 & 74.9 & 49.4 & 59.6 & 54.1 & 99.1 & 70.0 \\
\textsc{Single-Turn-Veri-Cls} & 75.0 & 77.6 & 31.1 & 69.0 & 74.0 & 49.3 & 59.2 & 54.3 & 98.9 & 70.1 \\
\textsc{Multi-Turn-SelfRef-Cls} & 72.7 & 75.3 & 33.0 & 67.3 & 72.1 & 49.1 & 58.4 & 53.6 & 99.0 & 69.5 \\
\midrule
\multicolumn{11}{c}{\textit{Qwen3-32B}} \\ 
\textsc{Two-Stage} & 73.6 & 75.0 & 22.6 & 71.7 & 60.4 & 51.4 & 55.6 & 58.9 & 94.1 & 72.5 \\
\textsc{Single-Turn} & 51.0 & 93.4 & 18.5 & 84.7 & 60.4 & 25.9 & 36.3 & 71.1 & 68.6 & 69.8 \\
\textsc{Single-Turn-Veri} & 97.8 & 99.2 & 31.8 & 88.5 & 65.9 & 31.2 & 42.4 & 71.5 & 74.0 & 72.7 \\
\textsc{Multi-Turn-SelfRef} & 91.0 & 95.0 & 40.9 & 89.7 & 69.5 & 40.7 & 51.3 & 68.4 & 79.3 & 73.5 \\
\textsc{Single-Turn-Cls} & 78.5 & 81.0 & 27.9 & 74.1 & 60.9 & 50.0 & 54.9 & 61.4 & 94.2 & 74.4 \\
\textsc{Single-Turn-Veri-Cls} & 78.8 & 81.0 & 29.9 & 73.3 & 61.5 & 51.7 & 56.2 & 61.6 & 93.8 & 74.3 \\
\textsc{Multi-Turn-SelfRef-Cls} & 86.1 & 89.9 & 41.0 & 83.9 & 71.7 & 51.5 & 59.9 & 65.8 & 92.0 & 76.7 \\
\bottomrule
\end{tabular}
\end{adjustbox}
\label{tab:full_main_result}
\end{table*}


\section{Qualitative Analysis}
\label{appendix:qualitative_analysis}

Through a manual review of incorrect model outputs, we identified several recurring error patterns that highlight the key challenges posed by our benchmark. The primary failure mode for each category is detailed below:

\begin{itemize}
    \item \texttt{answerable-correct}: The most common type of failure was an attempt to modify an already-correct candidate SQL query, which resulted in an incorrect version. It was infrequent for the model to misclassify these questions as not answerable. 
    \item \texttt{answerable-incorrect}: The most common failure was the inability to correct the provided incorrect SQL query.
    \item \texttt{ambiguous}: the most common failure was classifying the question as answerable and generating a SQL query without realizing the question's ambiguity. For example, when given the question is ``Albumin?'' and a candidate SQL provided an empty result, a model may decide to fix the query to not have an empty result rather than classifying the question as ambiguous. This could result in a final output like ``SELECT DISTINCT label FROM d\_labitems WHERE label = 'albumin'''. Misclassifying an ambiguous question as ``unanswerable'' was infrequent.
    \item \texttt{unanswerable}: Misclassifying the question as ``ambiguous'' was as common as misclassifying it as answerable and attempting to provide a SQL query. This was mainly due to models not taking the database schema (even though it was given as an input) into consideration when determining answerability. For example, for the question ``What was the name of the diagnosis for patient 10039997 in other departments?'', where no columns regarding departments exist, a model might misclassify the question as ambiguous because of the phrase ``other departments.'' This indicates the model failed to incorporate the provided knowledge of the database schema when classifying the question.
\end{itemize}


\section{Performance of Text-to-SQL Models for Candidate SQL Generation}
\label{appendix:sql_generation_performance}

\begin{table}[t!]
\centering
\footnotesize
\caption{Execution accuracy of seven different text-to-SQL models on MIMIC-IV, eICU, and MIMIC-III (MIMICSQL).}
\renewcommand{\arraystretch}{1.1}
\begin{adjustbox}{width=\linewidth,center}
\begin{tabular}{cccc}
\toprule
& \textbf{MIMIC-IV} & \textbf{eICU} & \textbf{MIMICSQL} \\
\midrule
\textsc{LLM-SQL} &  61.1 & 60.2 & 74.4 \\ 
\textsc{CodeS-15B} & 24.0 & 15.1 & 62.0 \\
\textsc{DIN-SQL} & 59.8 & 56.3 & 76.9 \\
\textsc{MAC-SQL} & 66.0 & 59.5 & 75.7 \\
\textsc{Deepseek R1-70B} & 46.7 & 54.2 & 76.0 \\
\textsc{Qwen3-32B} & \underline{69.3}  & \underline{62.3} & \textbf{86.9} \\
\textsc{OpenAI o4-mini} & \textbf{72.4}&\textbf{69.0} & \underline{85.9} \\
\midrule
\textsc{On Average} & 40.1 & 38.2 & 65.8 \\
\bottomrule

\end{tabular}
\end{adjustbox}
\vspace{-2mm}

\label{tab:sql_generator}
\end{table}

Table~\ref{tab:sql_generator} presents the performance comparison of seven text-to-SQL models across three medical datasets: MIMIC-IV, eICU, and MIMIC-III (MIMICSQL). Notably, OpenAI's o4-mini achieves the highest accuracy, with Qwen3-32B ranking second, while CodeS-15B yields the lowest performance.


\section{SQL Error Types and Definitions} \label{appendix:sql_error_types}
Table~\ref{appendix:sql_error_types_examples} presents the examples of error types present in the candidate SQL query from the answerable questions of the \textsc{sCARE} benchmark. In the table, T/C, J/G, PV, OL, and OG denote table/column reference errors, JOIN/GROUP BY errors, predicate value errors, other local errors, and other global errors, respectively.

\newcolumntype{E}{>{\centering\ttfamily}p{0.08\textwidth}}
\newcolumntype{F}{>{\RaggedRight}p{0.20\textwidth}}
\newcolumntype{G}{>{\RaggedRight\ttfamily}p{0.33\textwidth}}
\newcolumntype{H}{>{\RaggedRight\ttfamily}p{0.33\textwidth}}

\begin{table*}[t!]
\centering
\footnotesize
\renewcommand{\arraystretch}{1.0}
\caption{Examples of different SQL error types from the answerable questions in the \textsc{SCARE} benchmark. Each row shows a specific error in the candidate query: \textbf{T/C}, the query references a wrong column name in the SELECT clause; \textbf{J/G}, the query incorrectly joins the wrong table; \textbf{PV}, the query uses an incorrect value for the \texttt{long\_title} column; \textbf{OL}, the query unnecessarily uses a ROUND() function; \textbf{OG}, the query logic is too simplistic and misses the core user intent.}

\label{appendix:sql_error_types_examples}
\begin{tabular}{@{} E F G H @{}}
\toprule
\multicolumn{1}{c}{\textbf{Type}} & 
\multicolumn{1}{l}{\textbf{User Question}} & 
\multicolumn{1}{l}{\textbf{Ground-truth SQL}} & 
\multicolumn{1}{l}{\textbf{Candidate SQL}} \\
\midrule
\textbf{T/C} &
What was the name of the specimen test that was last given to patient 10027602 since 12/2100? &
SELECT microbiologyevents. spec\_type\_desc FROM microbiologyevents WHERE microbiologyevents.hadm\_id IN ( SELECT admissions.hadm\_id FROM admissions WHERE admissions.subject\_id = 10027602 ) AND strftime('\%Y-\%m',microbiologyevents.

charttime) >= '2100-12' ORDER BY microbiologyevents.charttime DESC LIMIT 1 &
SELECT test\_name FROM microbiologyevents WHERE subject\_id = 10027602 AND datetime(charttime) >= datetime('2100-12-31 23:59:00','start of month','-0 month') ORDER BY charttime DESC LIMIT 1 \\
\midrule
J/G &
What was patient 10015860's last value of anion gap in 10/this year? &
SELECT labevents.valuenum FROM labevents WHERE labevents.hadm\_id IN ( SELECT admissions.hadm\_id FROM admissions WHERE admissions.subject\_id = 10015860 ) AND labevents.itemid IN ( SELECT d\_labitems.itemid FROM d\_labitems WHERE d\_labitems.label = 'anion gap' ) AND datetime(labevents.charttime,'start of year') = datetime('2100-12-31 23:59:00','start of year','-0 year') AND strftime('\%m',labevents.charttime) = '10' ORDER BY labevents.charttime DESC LIMIT 1  &
SELECT ce.valuenum FROM chartevents ce JOIN d\_items di ON ce.itemid = di.itemid WHERE ce.subject\_id = 10015860 AND di.label = 'anion gap' AND datetime(ce.charttime,'start of year') = datetime('2100-12-31 23:59:00','start of year','-0 year') AND strftime('\%m', ce.charttime) = '10' ORDER BY ce.charttime DESC LIMIT 1 \\
\midrule
PV &
How many people were given procedure on single vessel two or more times? &
SELECT COUNT(T1.subject\_id) FROM ( SELECT admissions.subject\_id, COUNT(*) AS C1 FROM procedures\_icd JOIN admissions ON procedures\_icd.hadm\_id = admissions.hadm\_id WHERE procedures\_icd.icd\_code = ( SELECT d\_icd\_procedures.icd\_code FROM d\_icd\_procedures WHERE d\_icd\_procedures.long\_title = 'procedure on single vessel' ) GROUP BY admissions.subject\_id ) AS T1 WHERE T1.C1 >= 2 & SELECT COUNT(DISTINCT procedures\_icd.subject\_id) FROM procedures\_icd JOIN d\_icd\_procedures ON procedures\_icd.icd\_code = d\_icd\_procedures.icd\_code WHERE d\_icd\_procedures.long\_title = 'single vessel' GROUP BY procedures\_icd.subject\_id HAVING COUNT(*) >= 2
 \\
\midrule
\textbf{OL} &
What was the duration in days of patient 10018845's last stay in the icu? &
SELECT strftime('\%J', icustays.outtime) - strftime('\%J', icustays.intime) FROM icustays WHERE icustays.hadm\_id IN ( SELECT admissions.hadm\_id FROM admissions WHERE admissions.subject\_id = 10018845 ) AND icustays.outtime IS NOT NULL ORDER BY icustays.intime DESC LIMIT 1 &
SELECT ROUND((JULIANDAY(outtime) - JULIANDAY(intime)), 0) AS duration\_days FROM icustays WHERE subject\_id = 10018845 ORDER BY intime DESC LIMIT 1 \\
\midrule
\textbf{OG} &
Can you show me the top three most frequent lab tests in 2100? &
SELECT d\_labitems.label FROM d\_labitems WHERE d\_labitems.itemid IN ( SELECT T1.itemid FROM ( SELECT labevents.itemid, DENSE\_RANK() OVER ( ORDER BY COUNT(*) DESC ) AS C1 FROM labevents WHERE strftime('\%Y',labevents.charttime) = '2100' GROUP BY labevents.itemid ) AS T1 WHERE T1.C1 <= 3 ) & SELECT T1.label, COUNT(*) as frequency FROM d\_labitems T1 INNER JOIN labevents T2 ON T1.itemid = T2.itemid
\\
\bottomrule
\end{tabular}
\vspace{-2mm}
\end{table*}



\section{Details of Ambiguous and Unanswerable Question Annotation}
\label{apd:question_curation_annotation}

We provide further details on the generation and annotation process for the six categories of ambiguous and unanswerable questions introduced in Section~\ref{sec:ambig_unans_anno}. Table \ref{tab:sample_questions} shows samples for each category alongside an answerable example.

\begin{table*}[t!]
\centering
\huge
\renewcommand{\arraystretch}{1.3}
\caption{Examples of diverse question types in \textsc{SCARE}.}
\begin{adjustbox}{width=\linewidth,center}  
\begin{tabular}{ccc}
\toprule
\textbf{Type} & \multicolumn{1}{c}{\textbf{User Question}} & \textbf{Reason} \\
\hline
\texttt{answerable} & What are the five commonly ordered medications for patients aged 60 or above? & \makecell{Clear and answerable} \\
\midrule
\texttt{vague-question} & \textit{BP?} & \makecell{Too short, unclear intent} \\
\midrule
\texttt{vague-word} & How many current patients meet the \textit{high-risk} criteria? & \makecell{Term \textit{high-risk} not defined} \\
\midrule
\texttt{ambiguous-reference} & When was the first time \textit{it} happened in this hospital visit? & \makecell{Pronoun \textit{it} is ambiguous} \\
\midrule
\texttt{small talk} & \textit{Did you grab coffee?} & Not relevant to EHR data \\
\midrule
\texttt{out-of-scope} & Can you \textit{cluster} patients based on their medication patterns? & \makecell{Beyond text-to-SQL tasks} \\ 
\midrule
\texttt{missing-column} & What is the \textit{address} of patient 10016742? & \makecell{\textit{Address} not stored in MIMIC-IV/eICU} \\
\bottomrule
\end{tabular}
\end{adjustbox}
\vspace{-2mm}
\label{tab:sample_questions}
\end{table*}

\subsection{Ambiguous Questions}

The following details the generation process for the three types of ambiguous questions.

\begin{itemize}
    \item \textbf{\texttt{vague-question}}: We prompt GPT-4o to create overly vague, keyword-based questions \citep{radhakrishnan2020colloql} conditioned on a hospital domain, simulating user queries that lack specific intent (\textit{e.g.}, ``Weight?'', ``Symptoms?'').
    \item \textbf{\texttt{vague-word}}: To generate questions with imprecise filtering conditions, we prompt GPT-4o to strategically insert ambiguous terms (e.g., ``common,'' ``long,'' ``more than usual'') into otherwise answerable questions.
    \item \textbf{\texttt{ambiguous-reference}}: We create questions with unresolved references by prompting GPT-4o to modify answerable questions, incorporating referentially ambiguous words like ``this,'' ``that,'' or ``them.''
\end{itemize}

\subsection{Unanswerable Questions}

We create three types of unanswerable questions to test a system's ability to recognize queries beyond its scope.

\begin{itemize}
    \item \textbf{\texttt{small-talk}}: We use GPT-4o to generate conversational questions unrelated to EHR data (e.g., ``Did you grab coffee before rounds today?''), with instructions to explicitly avoid referencing the database schema.
    \item \textbf{\texttt{out-of-scope}}: We generate these by prompting GPT-4o to transform existing answerable questions into analytical tasks that extend beyond SQL's capabilities, such as predictive modeling.
    \item \textbf{\texttt{missing-column}}: In addition to using examples from EHRSQL, we generate a new set of more difficult questions for this category. We increase the difficulty by designing questions that reference columns that are plausible within the EHR context but do not exist in the database schema.
\end{itemize}

\paragraph{Quality Check} 
To ensure the reliability and consistency of our annotations, all generated questions undergo a rigorous review process. The process is conducted by three annotators (two authors and one hired external annotator), all of whom are computer science graduate students proficient in SQL. Each annotator independently evaluates whether the questions fit their designated categories. We measure inter-annotator agreement using Cohen's kappa, which ranges from 85.8\% to 90.9\%, and Fleiss' kappa, which is 87.8\%, indicating a high level of agreement. Only questions that receive unanimous approval from all three annotators are included in the final dataset.

\section{Baseline Method Implementation}
\label{appendix:prompt}

We present the prompts used to implement our four base methods: \textsc{Two-Stage}, \textsc{Single-Turn}, \textsc{Single-Turn-Veri}, and \textsc{Multi-Turn-SelfRef}. Note that the \{evidence\} part in the prompt refers to assumptions made during SQL annotation that are not explicitly stated in the questions (e.g., use SQLite for SQL query generation; use DENSE\_RANK() only when ranking is explicitly specified).

\subsection{\textsc{Two-Stage}}
\label{appendix:two_stage}

\vspace{1em} 

\begin{promptbox}[prompt:two-stage-classification]{Prompt used for the classification stage in \textsc{Two-Stage}.}
\lstset{
    basicstyle=\ttfamily\small,
    breaklines=true,
    breakindent=0pt,
    columns=fullflexible,
    upquote=true,
}

\begin{lstlisting}
Your task is to classify a natural language question into one of the following three categories, based on whether it can be answered using SQL over the given database schema.

Classification Categories:
1. **"answerable"** - The question can be clearly answered with the given database schema. All required information (tables, columns, relationships) exists in the schema.

2. **"ambiguous"** - The question is unclear, ambiguous, or requires clarification. This includes:
   - Questions with unclear references ("it", "that", "the previous one")
   - Questions with multiple possible interpretations without further clarification
   - Questions that are too vague to understand the intent

3. **"unanswerable"** - The question cannot be answered with the given database schema. This includes:
   - Questions requiring information not available in the database
   - Questions that are completely out of scope for SQL operations
   - Questions requiring functionality outside of SQL operations
   - Questions that are general conversation or small talk

# Input
- Database Schema
- Question

# Output Format
Respond with a single JSON object:
{{
  "reasoning": "<reasoning behind your decision>",
  "answer": "<one of the three categories: "answerable", "ambiguous", or "unanswerable">"
}}

# Input
Database Schema:
{database_schema}

Question: {question}
\end{lstlisting}
\end{promptbox}

\vspace{1em} 

\begin{promptbox}[prompt:two-stage-correction]{Prompt used for the SQL correction stage in \textsc{Two-Stage}.}
\lstset{
    basicstyle=\ttfamily\small,
    breaklines=true,
    breakindent=0pt,
    columns=fullflexible,
    upquote=true,
}

\begin{lstlisting}
Your task is to check whether the given SQL query is correct according to the schema, the question, and the SQL guideline.
If it is incorrect, provide a corrected query. If it is correct, return it unchanged.

# Input
- Database Schema
- Question
- Evidence (SQL guideline)
- Current SQL Query
- Execution Result

# Output Format
Respond with a single JSON object:
{{
  "reasoning": "<reasoning behind your decision>",
  "answer": "the corrected SQL query if incorrect, otherwise the original query"
}}

# Input
Database Schema:
{database_schema}

Question:
{question}

Evidence:
{evidence}

Current SQL Query:
{sql}

Execution Result:
{exec}
\end{lstlisting}
\end{promptbox}


\subsection{\textsc{Single-Turn}}
\label{appendix:single_turn}

\vspace{1em} 

\begin{promptbox}[prompt:single-turn]{The prompt used in \textsc{Single-Turn}.}
\lstset{
    basicstyle=\ttfamily\small,
    breaklines=true,
    breakindent=0pt,
    columns=fullflexible,
    upquote=true,
}

\begin{lstlisting}
Your task is to determine whether the predicted SQL is correct, or whether the question is intrinsically ambiguous or unanswerable for SQL generation, given the question and database schema. Follow the instructions below:

- If the question is **answerable** and the SQL is **correct**, output the same SQL.
- If the question is **answerable** but the SQL is **incorrect**, output the fixed SQL.
- If the question is **ambiguous** (requires clarification before SQL translation), output `"ambiguous"`. Ambiguity types include:
  1. vague-question: Extremely brief or unclear questions (e.g., "BP?", "Patient?").
  2. vague-word: Questions containing vague words (e.g., "How many patients meet the *high risk* criteria?").
  3. referential-ambiguity: Questions containing unclear referents (e.g., "When was the first time *it* happened in the most recent hospital visit?").
- If the question is **unanswerable** (cannot be converted into valid SQL), output `"unanswerable"`. Unanswerable types include:
  1. small-talk: Casual queries unrelated to the data (e.g., "Did you grab coffee?").
  2. out-of-scope: Requests that cannot be handled by SQL (e.g., "Can you *cluster* patients based on medication patterns?").
  3. missing-column: References to non-existent columns (e.g., asking for an *address* field not present in the schema).

# Important
- Do not assume anything that is not explicitly stated in the input.

# Input
- Database Schema
- Question
- Evidence (SQL guideline)
- Predicted SQL
- Execution Result

# Output Format
Respond with a single JSON object:
{{
  "reasoning": "<reasoning behind your decision>",
  "answer": "<either the original SQL, fixed SQL, "ambiguous", or "unanswerable">"
}}

# Input
Database Schema:
{database_schema}

Question:
{question}

Evidence:
{evidence}

Predicted SQL:
{sql}

Execution Result:
{exec}
\end{lstlisting}
\end{promptbox}


\subsection{\textsc{Single-Turn-Veri}}
\label{appendix:single_turn_veri}

\vspace{1em} 

\begin{promptbox}[prompt:single-turn-veri-correct]{The SQL correction prompt used in \textsc{Single-Turn-Veri}.}
\lstset{
    basicstyle=\ttfamily\small,
    breaklines=true,
    breakindent=0pt,
    columns=fullflexible,
    upquote=true,
}

\begin{lstlisting}
Your task is to determine whether the predicted SQL is correct, or whether the question is intrinsically ambiguous or unanswerable for SQL generation, given the question and database schema. Follow the instructions below:

- If the question is **answerable** and the SQL is **correct**, output the same SQL.
- If the question is **answerable** but the SQL is **incorrect**, output the fixed SQL.
- If the question is **ambiguous** (requires clarification before SQL translation), output `"ambiguous"`. Ambiguity types include:
  1. vague-question: Extremely brief or unclear questions (e.g., "BP?", "Patient?").
  2. vague-word: Questions containing vague words (e.g., "How many patients meet the *high risk* criteria?").
  3. referential-ambiguity: Questions containing unclear referents (e.g., "When was the first time *it* happened in the most recent hospital visit?").
- If the question is **unanswerable** (cannot be converted into valid SQL), output `"unanswerable"`. Unanswerable types include:
  1. small-talk: Casual queries unrelated to the data (e.g., "Did you grab coffee?").
  2. out-of-scope: Requests that cannot be handled by SQL (e.g., "Can you *cluster* patients based on medication patterns?").
  3. missing-column: References to non-existent columns (e.g., asking for an *address* field not present in the schema).

# Important
- Do not assume anything that is not explicitly stated in the input.

# Input
- Database Schema
- Question
- Evidence (SQL guideline)
- Predicted SQL
- Execution Result

# Output Format
Respond with a single JSON object:
{{
  "reasoning": "<reasoning behind your decision>",
  "answer": "<either the original SQL, fixed SQL, "ambiguous", or "unanswerable">"
}}

# Input
Database Schema:
{database_schema}

Question:
{question}

Evidence:
{evidence}

Predicted SQL:
{sql}

Execution Result:
{exec}
\end{lstlisting}
\end{promptbox}




\vspace{1em} 

\begin{promptbox}[prompt:single-turn-veri-veri]{The verifier prompt used in \textsc{Single-Turn-Veri}.}
\lstset{
    basicstyle=\ttfamily\small,
    breaklines=true,
    breakindent=0pt,
    columns=fullflexible,
    upquote=true,
}
\begin{lstlisting}
Your task is to verify whether the model has correctly followed the task instructions for SQL prediction. Carefully evaluate the predicted SQL in relation to the database schema, the question, the evidence (SQL guideline), and the execution result.

Follow the instructions below:
- If the predicted SQL is correct (or if the question is not answerable and the label is "ambiguous" or "unanswerable"), start your feedback with the phrase "the predicted SQL is correct".
- If the predicted SQL is incorrect, start your feedback with the phrase "the predicted SQL is incorrect". Then explain clearly why it is wrong (e.g., incorrect column, wrong join, missing condition, misclassification of ambiguity/unanswerability, etc.).

Ambiguity Types
  1. vague-question: Extremely brief or unclear questions (e.g., "BP?", "Patient?").
  2. vague-word: Questions containing vague words (e.g., "How many patients meet the *high risk* criteria?").
  3. referential-ambiguity: Questions containing unclear referents (e.g., "When was the first time *it* happened in the most recent hospital visit?").  

Unanswerable Types
  1. small-talk: Casual queries unrelated to the data (e.g., "Did you grab coffee?").
  2. out-of-scope: Requests that cannot be handled by SQL (e.g., "Can you *cluster* patients based on medication patterns?").
  3. missing-column: References to non-existent columns (e.g., asking for an *address* field not present in the schema).

# Important
- Your role is to provide evaluation feedback only, not to generate or fix SQL
- Your feedback should be precise and grounded in the given schema and instruction.
- Do not assume anything that is not explicitly stated in the input.

# Input
- Database Schema
- Question
- Question Explanation
- Evidence (SQL guideline)
- Predicted SQL
- SQL Explanation
- Execution Result

# Output Format
Respond with a single JSON object:
{{
  "feedback": "<detailed feedback on your decision>"
}}

# Input
Database Schema:
{database_schema}

Question:
{question}

Evidence:
{evidence}

Predicted SQL:
{sql}

Execution Result:
{exec}
\end{lstlisting}
\end{promptbox}


\subsection{\textsc{Multi-Turn-SelfRef}}
\label{appendix:multi_turn_Selfref}

\vspace{1em} 

\begin{promptbox}[prompt:multi-turn-selfref]{Prompt used for the SQL correction stage in \textsc{Multi-Turn-SelfRef}.}
\lstset{
    basicstyle=\ttfamily\small,
    breaklines=true,
    breakindent=0pt,
    columns=fullflexible,
    upquote=true,
}
\begin{lstlisting}
Your task is to determine whether the predicted SQL is correct, or whether the question is intrinsically ambiguous or unanswerable for SQL generation, given the question and database schema. Follow the instructions below:

- If the question is **answerable** and the SQL is **correct**, output the same SQL.
- If the question is **answerable** but the SQL is **incorrect**, output the fixed SQL.
- If the question is **ambiguous** (requires clarification before SQL translation), output `"ambiguous"`. Ambiguity types include:
  1. vague-question: Extremely brief or unclear questions (e.g., "BP?", "Patient?").
  2. vague-word: Questions containing vague words (e.g., "How many patients meet the *high risk* criteria?").
  3. referential-ambiguity: Questions containing unclear referents (e.g., "When was the first time *it* happened in the most recent hospital visit?").
- If the question is **unanswerable** (cannot be converted into valid SQL), output `"unanswerable"`. Unanswerable types include:
  1. small-talk: Casual queries unrelated to the data (e.g., "Did you grab coffee?").
  2. out-of-scope: Requests that cannot be handled by SQL (e.g., "Can you *cluster* patients based on medication patterns?").
  3. missing-column: References to non-existent columns (e.g., asking for an *address* field not present in the schema).

# Important
- Do not assume anything that is not explicitly stated in the input.

# Input
- Database Schema
- Question
- Evidence (SQL guideline)
- Predicted SQL
- Execution Result

# Output Format
Respond with a single JSON object:
{{
  "reasoning": "<reasoning behind your decision>",
  "answer": "<either the original SQL, fixed SQL, "ambiguous", or "unanswerable">"
}}

# Input
Database Schema:
{database_schema}

Question:
{question}

Evidence:
{evidence}

Predicted SQL:
{sql}

Execution Result:
{exec}
\end{lstlisting}
\end{promptbox}

\vspace{1em} 

\begin{promptbox}[prompt:multi-turn-selfref-feedback]{Prompt used for the feedback stage in \textsc{Multi-Turn-SelfRef}.}
\lstset{
    basicstyle=\ttfamily\small,
    breaklines=true,
    breakindent=0pt,
    columns=fullflexible,
    upquote=true,
}

\begin{lstlisting}
Your task is to review whether the model has followed the task instructions correctly for SQL prediction. Carefully examine the database schema, the question, the evidence (SQL guideline), the predicted SQL, and the execution result.

Follow the instructions below:
- If the predicted SQL is correct (or if the question is not answerable and the label is "ambiguous" or "unanswerable"), start your feedback with the phrase "the predicted SQL is correct".
- If the predicted SQL is incorrect, start your feedback with the phrase "the predicted SQL is incorrect". Then explain clearly why it is wrong (e.g., incorrect column, wrong join, missing condition, misclassification of ambiguity/unanswerability, etc.).

Ambiguity Types
  1. vague-question: Extremely brief or unclear questions (e.g., "BP?", "Patient?").
  2. vague-word: Questions containing vague words (e.g., "How many patients meet the *high risk* criteria?").
  3. referential-ambiguity: Questions containing unclear referents (e.g., "When was the first time *it* happened in the most recent hospital visit?").  

Unanswerable Types
  1. small-talk: Casual queries unrelated to the data (e.g., "Did you grab coffee?").
  2. out-of-scope: Requests that cannot be handled by SQL (e.g., "Can you *cluster* patients based on medication patterns?").
  3. missing-column: References to non-existent columns (e.g., asking for an *address* field not present in the schema).

# Important
- Your role is to provide evaluation feedback only, not to generate or fix SQL
- Your feedback should be precise, grounded in the given schema and instruction.
- Do not assume anything that is not explicitly stated in the input.

# Input
- Database Schema
- Question
- Question Explanation
- Evidence (SQL guideline)
- Predicted SQL
- SQL Explanation
- Execution Result

# Response Format
{{
  "feedback": "<detailed feedback on your decision>"
}}

# Input
Database Schema:
{database_schema}

Question:
{question}

Evidence:
{evidence}

Predicted SQL:
{sql}

Execution Result:
{exec}
\end{lstlisting}
\end{promptbox}

\vspace{1em} 

\begin{promptbox}[prompt:multi-turn-serlfref-refine]{Prompt used for the refinement stage in \textsc{Multi-Turn-SelfRef}.}
\lstset{
    basicstyle=\ttfamily\small,
    breaklines=true,
    breakindent=0pt,
    columns=fullflexible,
    upquote=true,
}
\begin{lstlisting}
All previous SQL queries were found to be problematic. Based on the feedback, determine once again whether the question is answerable, ambiguous, or unanswerable. If it is answerable, correct the SQL.

Predicted SQL:
{sql}

Execution Result:
{exec}
\end{lstlisting}
\end{promptbox}

\end{document}